\documentclass[preprint,11pt,authoryear]{elsarticle}

\usepackage{amssymb}
\usepackage{amsmath}

\usepackage{amssymb}
\usepackage{amsmath}
\usepackage{color}
\usepackage{float}
\usepackage{booktabs, multirow}
\usepackage{arydshln}  
\usepackage{makecell}   

\usepackage{amsmath, amssymb}
\usepackage{algorithm}
\usepackage{algorithmic} 
\usepackage{float}       
\usepackage{caption}     

\usepackage{amsthm}
\newtheorem{proposition}{Proposition}[section]
\newtheorem{corollary}{Corollary}[proposition]

\usepackage{algorithm}
\usepackage{algorithmic}
\usepackage{float}

\usepackage{enumitem}

\usepackage{lipsum}

\usepackage{booktabs}
\usepackage[table]{xcolor}
\usepackage{caption}
\usepackage{geometry}
\usepackage{svg}
\usepackage{graphicx}
\usepackage{amsthm}

\usepackage{amsthm}
\theoremstyle{definition}

\usepackage{setspace}
\journal{European Journal of Operational Research}

\begin{document}
\begin{frontmatter}

\title{A Robust and Non-Iterative Tensor Decomposition Method with Automatic Thresholding}

\author[1]{Hiroki Hasegawa}
\affiliation[1]{organization={Graduate School of Science and Technology, University of Tsukuba},
            addressline={1-1-1 Tennodai}, 
            city={Tsukuba},
            postcode={305-8573}, 
            state={Ibaraki},
            country={Japan}}

\author[2,3,4]{Yukihiko Okada}
\affiliation[2]{organization={Institute of Systems and Information Engineering, University of Tsukuba},
            addressline={1-1-1 Tennodai}, 
            city={Tsukuba},
            postcode={305-8577}, 
            state={Ibaraki},
            country={Japan}}
            
\affiliation[3]{organization={Tsukuba Institute for Advanced Research, University of Tsukuba},
            addressline={1-1-1 Tennodai}, 
            city={Tsukuba},
            postcode={305-8577}, 
            state={Ibaraki},
            country={Japan}}
            
\affiliation[4]{organization={Center for Artificial Intelligence Research, University of Tsukuba},
            addressline={1-1-1 Tennodai}, 
            city={Tsukuba},
            postcode={305-8577}, 
            state={Ibaraki},
            country={Japan}}

\begin{abstract}
Recent progress in IoT and biometric sensing technologies has produced massive, high-dimensional tensor data, but achieving accurate and efficient low-rank approximation remains challenging. Most existing tensor decomposition methods require predefined ranks and iterative optimization, leading to high computational costs and reliance on analyst expertise. This study proposes a new tensor low-rank approximation method that eliminates both prior rank specification and iterative optimization. The method applies statistical singular value hard thresholding to each mode-wise unfolded matrix to automatically extract statistically significant components, reducing noise while preserving intrinsic structure. Theoretically, optimal thresholds for each mode are derived from the asymptotic properties of the Marčenko–Pastur distribution. Simulations show that the proposed method outperforms conventional approaches—HOSVD, HOOI, and Tucker-L2E—in both estimation accuracy and computational efficiency. These results indicate that the method offers a theoretically grounded, fully automatic, and non-iterative framework for tensor decomposition.
\end{abstract}

\begin{keyword}
Multivariate statistics \sep Tensor decomposition \sep Automatic thresholding \sep Noise reduction \sep Data science
\end{keyword}

\end{frontmatter}

\section{Introduction}
Vast amounts of data are continuously collected from diverse sources, including sensors, social networks, and biometric systems. Such data often exhibit tensor structures capturing temporal, spatial, and categorical relationships, forming a foundation for multi-perspective analysis\citep{Kolda2009}. However, real-world data collection involves measurement errors and environmental fluctuations, inherently introducing noise. Such noise obscures meaningful patterns and undermines analytical reliability \citep{Haven2012}.

In Operations Research (OR), the importance of real-world data analysis has been steadily increasing. Yet, noise remains inevitable due to measurement and environmental limitations, making its proper treatment a key issue\citep{Haven2012}. Prior studies have proposed various approaches to mitigate noise effects, including feature selection for improved classification accuracy \citep{Piramuthu2004}, the proposal of dynamic resampling strategies in evolutionary optimization \citep{Syberfeldt2010}, the improvement of price forecasting accuracy through wavelet transforms \citep{Haven2012}, and the construction of models capable of estimating both short-term and long-term inefficiencies \citep{Badunenko2016}.

While all these studies sought to suppress noise and improve accuracy, most relied on matrix-based frameworks. However, modern data environments increasingly involve high-dimensional data, such as sensor or biometric signals that intertwine temporal, spatial, and categorical elements. In such cases, it is difficult for matrix-based analytical methods to avoid the loss of structural information. In particular, as the number of variables and observation points increases, the so-called “curse of dimensionality” becomes apparent, leading to an increase in computational burden and the risk of overlooking latent structures \citep{Kilmer2019, Auddy2024}.

To address these challenges, tensor analysis has gained attention for its ability to preserve multidimensional data structures during analysis. Because tensors can handle high-dimensional data while maintaining relationships among different modes, they are particularly effective in applications that simultaneously consider multiple dimensions such as time, space, and category (e.g., demand forecasting and anomaly detection) \citep{Kolda2009}. Furthermore, it has been shown that low-rank approximation can suppress the influence of noise and missing data, thereby achieving more accurate predictions and robust anomaly detection \citep{Acar2011, ZhangAeron2016}. Against this background, a promising strategy has been positioned as one that separates observed data into signal and noise components and removes noise by estimating the signal. In particular, tensor decomposition methods such as CP decomposition, Tucker decomposition, and Tensor-Train decomposition have been attracting attention as powerful tools for noise removal and structural recovery in multidimensional data \citep{Carroll1970, Harshman1970, Tucker1966, Oseledets2011}.

The first category of approaches is based on CANDECOMP/PARAFAC (CP) decomposition, including methods such as CP-ALS, Bayesian Robust Tensor Factorization (BRTF), Randomized Block Sampling-CPD, Tensor Robust CP Decomposition, and Randomized CP Decomposition \citep{Kolda2009, Zhao2015, Vervliet2015, Xue2017, Erichson2020}. CP-ALS is the canonical method that iteratively updates factor matrices via alternating least squares. BRTF leverages Bayesian inference to robustly estimate low-rank structures while suppressing the influence of outliers. Randomized Block Sampling-CPD improves computational efficiency on large-scale tensors by randomly sampling tensor blocks. Tensor Robust CP Decomposition enables accurate extraction of low-rank structure even in the presence of sparse outliers. Randomized CP Decomposition employs randomized algorithms to efficiently approximate CP decompositions.

The second category centers on Tucker decomposition, encompassing methods such as HOSVD, HOOI, HoRPCA-W, HoRPCA-C, HoRPCA-S, TORCA, Zoom-Tucker, and Tucker-L2E \citep{DeLathauwer2000a, DeLathauwer2000b, Tomioka2011, Goldfarb2014, Xue2017, Jang2021, Heng2023}. As a higher-order extension of principal component analysis, Tucker decomposition provides significant modeling flexibility, leading to numerous variants \citep{Kolda2009}. HOSVD performs singular value decomposition on each tensor mode to extract features while preserving multidimensional structure. HOOI improves upon HOSVD by iteratively refining factor matrices for better approximation accuracy. HoRPCA-W introduces a weighted loss function to decompose a tensor into low-rank, noise, and sparse components in a robust manner. HoRPCA-C adopts a Cauchy loss to further reduce sensitivity to outliers, while HoRPCA-S applies smoothed L1 regularization to achieve stable and smooth separation. TORCA enforces orthogonality in factor matrices while maintaining robustness against outliers. Zoom-Tucker balances local decomposition accuracy and computational efficiency by restricting optimization to temporal regions. Tucker-L2E enhances robustness to outliers through L2-based loss optimization, offering stable performance even on higher-rank data.

The third category involves Tensor-Train (TT) decomposition, a more recent approach compared to CP and Tucker, with growing interest but limited application scope to date. Representative methods include TTPUDR, GRTT Decomposition, FastTT, TTOI, and TT-ICE \citep{Bai2019, Sofuoglu2020, Li2022, Zhou2022, Aksoy2024}. TTPUDR combines TT decomposition with locality-preserving projections for structure-aware dimensionality reduction of high-dimensional data. GRTT Decomposition integrates graph regularization to preserve local structures and learn low-rank TT representations. FastTT is designed for fast and accurate decomposition of large-scale and sparse tensors. TTOI sequentially estimates TT structures via alternating forward and inverse singular value decompositions with projections, achieving high accuracy even under noise. TT-ICE provides guaranteed precision in sequentially updating TT-format approximations, simultaneously attaining high compression rates and computational efficiency.

Although the aforementioned CP, Tucker, and TT decomposition methods were developed to achieve low-rank approximations of tensors, 
most of them require a priori specification of rank parameters and rely on iterative algorithms for optimizing their objective functions. 
In particular, estimating the rank of a tensor is known to be NP-hard, and identifying an optimal rank configuration often entails substantial computational cost \citep{Hillar2013}. 
Consequently, many existing approaches adopt fixed-rank assumptions and employ iterative optimization schemes, which can pose practical limitations. In the field of OR, similar limitations have been observed in tensor-based analytical models. 
\citet{Chen2016} proposed the Multi-Kernel Support Tensor Machine (MK-STM), which integrates heterogeneous multiway data through a dual SVM formulation and linear programming for kernel weight estimation, requiring iterative optimization. \citet{Chen2020} developed the Collaborative Multiway Data Factorization (CMDF), which jointly decomposes multiple tensors using an alternating least squares (ALS) scheme similar to the Higher-Order Orthogonal Iteration (HOOI). 
Although both studies demonstrated the effectiveness of tensor representations in marketing analytics and customer modeling, they still depend on iterative optimization and manually specified parameters, indicating the need for a non-iterative and rank-free framework. As shown in Table~\ref{tab:tucker-comparison}, such constraints may hinder the applicability and scalability of these methods in real-world settings.

\begin{table*}[h]
\centering
\caption{Comparison of Tensor Decomposition Methods}
\label{tab:tucker-comparison}
\begin{tabular}{@{}p{3.5cm}clccc@{}}
\toprule
\textbf{Method Name} & \textbf{Fundamentals} & \textbf{1st Author(Year)} & & \textbf{Estimation?} & \textbf{Iterative?} \\
\midrule
\makecell[l]{CP--ALS} & CP & Kolda (2009) & & Yes & Yes \\
\makecell[l]{BRTF} & CP & Zhao (2015) & & No & Yes \\
\makecell[l]{Randomized Block\\Sampling CPD} & CP & Vervliet (2015) & & Yes & Yes \\
\makecell[l]{Tensor Robust\\CP Decomposition} & CP & Xue (2017) & & Yes & Yes \\
\makecell[l]{Randomized\\CP Decomposition} & CP & Erichson (2020) & & Yes & Yes \\
\\[1ex]
HOSVD & Tucker & De Lathauwer (2000a) & & Yes & No \\
HOOI & Tucker & De Lathauwer (2000b) & & Yes & Yes \\
HoRPCA--W & Tucker & Tomioka (2011) & & No & Yes \\
HoRPCA--C & Tucker & Goldfarb (2014) & & Yes & Yes \\
HoRPCA--S & Tucker & Goldfarb (2014) & & No & Yes \\
TORCA & Tucker & Xue (2017) & & Yes & Yes \\
\makecell[l]{Zoom--Tucker} & Tucker & Jang (2021) & & Yes & Yes \\
\makecell[l]{Tucker--L2E} & Tucker & Heng (2023) & & No & Yes \\
\textbf{\makecell[l]{TARST}} & \textbf{Tucker} & \textbf{Hasegawa (2025)} && \textbf{No} & \textbf{No} \\
\\[1ex]
\makecell[l]{TTPUDR} & TT & Bai (2019) & & Yes & Yes \\
\makecell[l]{GRTT\\Decomposition} & TT & Sofuoglu (2020) & & Yes & Yes \\
\makecell[l]{FastTT} & TT & Li (2020) & & Yes & Yes \\
\makecell[l]{TTOI} & TT & Zhou (2022) & & Yes & Yes \\
\makecell[l]{TT--ICE} & TT & Aksoy (2024) & & No & Yes \\
\bottomrule
\end{tabular}

\vspace{2mm}
\footnotesize{
\textit{Note.} CP = Canonical Polyadic; Tucker = Tucker decomposition; TT = Tensor Train; 
“TARST” (Tensor Automatic Rank-free Singular-value Thresholding) denotes the non-iterative, rank-free Tucker decomposition method proposed in this study.
}
\end{table*}

The central research question of this study is whether it is possible to efficiently and reliably extract meaningful information from high-dimensional and noisy data using a tensor decomposition method that is both non-iterative and does not require predefined rank specification. To address this question, this study proposes a novel method, Tensor Automatic Rank-free Singular-value Thresholding (TARST). TARST is a non-iterative algorithm designed within the framework of Tucker-based higher-order singular value decomposition. The proposed method applies the matrix-based Singular value hard thresholding technique introduced by \citet{Gavish2014} to each mode unfolding of a tensor, automatically extracting only the significant components. This enables simultaneous noise removal and low-rank approximation while preserving the intrinsic tensor structure.

In particular, by setting thresholds based on the noise level of each mode, TARST statistically estimates the Tucker rank without requiring manual adjustment. As a result, it can avoid the NP-hard problem of rank selection. Furthermore, by extending the theoretical guarantees of \citet{Gavish2014} from matrices to tensors, TARST provides consistent and stable estimation without dependence on initialization or iterative optimization. Moreover, as TARST does not require the specification of any hyperparameters, it remains independent of the analyst's experience or technical expertise and thus ensures highly reproducible analysis.

\section{Methodology}
This section introduces the foundational components leading to our proposed method: (1) notation, (2) Higher-Order Singular Value Decomposition (HOSVD), (3) Singular value hard thresholding(SVHT), and (4) the formulation of the proposed method.

\subsection{Notation}
Throughout this paper, tensors are denoted by calligraphic uppercase letters (e.g., $\mathcal{X}, \mathcal{G}$), matrices by bold uppercase letters (e.g., $\mathbf{U}, \mathbf{V}$), vectors by bold lowercase letters (e.g., $\mathbf{u}, \mathbf{v}$), and scalars by Roman or Greek characters (e.g., $x, y, \lambda, \sigma$).
For an $N$-way tensor $\mathcal{X} \in \mathbb{R}^{I_1 \times \cdots \times I_N}$, its mode-$n$ unfolding is denoted as $\mathbf{X}{(n)} \in \mathbb{R}^{I_n \times (I_1 \cdots I{n-1} I_{n+1} \cdots I_N)}$, where mode-$n$ indices form the rows and the others are flattened into columns.

\subsection{HOSVD}

The HOSVD is a tensor decomposition method that builds upon the theoretical foundation of the Tucker decomposition. Tucker decomposition factorizes a given tensor $\mathcal{X} \in \mathbb{R}^{I_1 \times \cdots \times I_N}$ into a core tensor and a set of factor matrices as
\[
\mathcal{X} \approx \mathcal{G} \times_1 \mathbf{U}^{(1)} \times_2 \cdots \times_N \mathbf{U}^{(N)},
\]
where $\mathcal{G} \in \mathbb{R}^{R_1 \times \cdots \times R_N}$ is the core tensor that contains compressed information, and $\mathbf{U}^{(n)} \in \mathbb{R}^{I_n \times R_n}$ are the factor matrices corresponding to each mode\citep{Tucker1966}. 
The $n$-mode product of a tensor $\mathcal{A} \in \mathbb{R}^{I_1 \times \cdots \times I_N}$ and a matrix $\mathbf{U}^{(n)} \in \mathbb{R}^{J \times I_n}$ is defined element-wise as
\[
\left(\mathcal{A} \times_n \mathbf{U}^{(n)}\right)_{i_1, \dots, j, \dots, i_N}
= \sum_{i_n=1}^{I_n} \mathbf{U}^{(n)}_{j, i_n}\, \mathcal{A}_{i_1, \dots, i_n, \dots, i_N}.
\]

HOSVD can be viewed as a special case of the Tucker decomposition, in which each factor matrix $\mathbf{U}^{(n)}$ is constrained to be orthogonal and is obtained by applying the singular value decomposition (SVD) to the mode-$n$ unfolding matrix $\mathbf{X}_{(n)}$ of $\mathcal{X}$. 
The decomposition is given by
\[
\mathcal{X} = \mathcal{G} \times_1 \mathbf{U}^{(1)} \times_2 \cdots \times_N \mathbf{U}^{(N)},
\]
and the core tensor is computed as
\[
\mathcal{G} = \mathcal{X} \times_1 \left(\mathbf{U}^{(1)}\right)^\top \times_2 \cdots \times_N \left(\mathbf{U}^{(N)}\right)^\top.
\]
This orthogonality constraint enables stable and interpretable factorization; however, because the SVD is applied independently to each mode, HOSVD does not necessarily achieve the optimal compression that considers interactions across modes \cite{DeLathauwer2000a}. 
This limitation has motivated the development of iterative refinement methods such as the Higher-Order Orthogonal Iteration (HOOI)\citep{DeLathauwer2000b}.

\subsection{Singular Value Hard Thresholding(SVHT)}

SVHT is a matrix denoising technique that reconstructs a low-rank approximation by applying a nonlinear shrinkage function to the singular values of a noisy observed matrix. Given an observed matrix
\[
\mathbf{Y} = \mathbf{X} + \sigma \mathbf{Z},
\]
where $\mathbf{X}$ is the unknown low-rank signal and $\mathbf{Z}$ is a noise matrix whose entries are independent and independent and identically distributed (i.i.d.) random variables with zero mean and unit variance, i.e., $Z_{ij} \sim \mathcal{N}(0,1)$. Following \cite{Gavish2014}, the scalar $\sigma > 0$ represents the noise level, defined as the standard deviation of each additive noise entry. It controls the overall scale of the perturbation applied to the true signal $\mathbf{X}$, and therefore determines the magnitude of the singular value spread induced by noise.

SVHT operates on the SVD of $\mathbf{Y}$ as follows:
\[
\mathbf{Y} = \sum_{i=1}^{m} y_i \mathbf{u}_i \mathbf{v}_i^{\top}.
\]
Then, it estimates $\mathbf{X}$ by shrinking or discarding the singular values according to a chosen shrinkage rule $\eta(y_i; \tau)$:
\[
\hat{\mathbf{X}} = \sum_{i=1}^{m} \eta(y_i; \tau)\, \mathbf{u}_i \mathbf{v}_i^{\top}.
\]
In other words, SVD decomposes the data into orthogonal bases weighted by their singular values, and SVHT modifies these weights to suppress noise.

A simple and widely used rule is the hard thresholding rule, which sets singular values smaller than a threshold $\tau$ to zero:
\[
\eta_H(y_i; \tau) =
\begin{cases}
y_i, & y_i \ge \tau, \\
0,   & y_i < \tau.
\end{cases}
\]
This process keeps only the significant components of the data while removing small, noise-dominated singular values.

In their influential paper, \cite{Gavish2014} theoretically determined the optimal threshold for this hard singular value truncation under an asymptotic framework where the matrix dimension $n$ grows large while the true rank remains fixed and small.
Specifically, for an $n \times n$ matrix corrupted by white Gaussian noise with variance $\sigma^2$, they proved that the asymptotically optimal threshold is
\[
\tau^* = \frac{4}{\sqrt{3}} \sqrt{n}\, \sigma \approx 2.309 \sqrt{n}\, \sigma.
\]
Here, ``asymptotically optimal'' means that as the matrix dimension increases, this threshold minimizes the asymptotic mean squared error (AMSE) between the true signal and its estimate.

For rectangular matrices, the threshold depends on the aspect ratio
\[
\beta = \frac{m}{n}, \quad (0 < \beta \le 1),
\]
which represents the ratio of the number of rows $m$ to columns $n$.
The optimal threshold generalizes to
\[
\tau^* = \lambda^*(\beta) \sqrt{n}\, \sigma,
\]
where $\lambda^*(\beta)$ is a coefficient that depends on the matrix shape. The optimal threshold values follow those derived by \cite{Gavish2014}. For example, when $\beta = 1$ (square matrix), $\lambda^*(1) = 4/\sqrt{3}$; as $\beta$ decreases, the optimal threshold becomes slightly smaller because the noise spectrum narrows for more rectangular matrices.

This rule—known as Optimal SVHT—minimizes the AMSE among all possible hard thresholds. The AMSE represents the large-sample limit of the expected reconstruction error:
\[
\mathrm{AMSE} = \lim_{n \to \infty} \mathbb{E}\left[\|\hat{\mathbf{X}}_n - \mathbf{X}_n\|_F^2\right],
\]
whereas for finite-sized matrices, the mean squared error (MSE) is defined as
\[
\mathrm{MSE} = \mathbb{E}\left[\|\hat{\mathbf{X}} - \mathbf{X}\|_F^2\right].
\]
While MSE depends on the specific realization of the noise and matrix size, AMSE provides a stable analytical measure of performance in the high-dimensional limit. Minimizing AMSE thus identifies the shrinkage rule that achieves the smallest average denoising error when both matrix dimensions grow large. In contrast, the traditional Truncated Singular Value Decomposition (Truncated SVD) reconstructs a low-rank approximation by retaining only the top-$r$ singular values and their corresponding singular vectors in the SVD of $\mathbf{Y}$, that is,
\[
\mathbf{Y}_r = \sum_{i=1}^{r} y_i \mathbf{u}_i \mathbf{v}_i^{\top},
\]
where $r$ denotes the target rank to be preserved. This operation yields the best rank-$r$ approximation of $\mathbf{Y}$ in the Frobenius norm, according to the Eckart--Young--Mirsky theorem~\citep{Eckart1936}. However, it requires prior knowledge of the true rank $r$, which is generally unknown in practice. Interestingly, even if the true rank $r$ were known, \cite{Gavish2014} demonstrated that SVHT achieves a smaller MSE than this truncated SVD, highlighting the theoretical advantage of data-driven thresholding over fixed-rank truncation.

Interestingly, even if the true rank $r$ were known, \cite{Gavish2014} showed that SVHT can achieve a smaller mean squared error than this truncated SVD, highlighting the theoretical advantage of data-driven thresholding over fixed-rank truncation.

When the noise level $\sigma$ is unknown, \cite{Gavish2014} also proposed a fully data-driven adaptive version, where $\sigma$ is estimated from the data itself using the median of the observed singular values. This approach exploits the fact that, for a pure noise matrix, the singular values follow the Marčenko-Pastur (MP) distribution.
The median singular value $y_{\mathrm{med}}$ therefore provides a robust estimate of the noise scale:
\[
\hat{\sigma}(Y) = \frac{y_{\mathrm{med}}}{\sqrt{n\,\mu_\beta}},
\]
where $\mu_\beta$ is the median of the MP distribution with aspect ratio $\beta = m/n$. Substituting this estimate into the optimal threshold $\tau^* = \lambda^*(\beta)\sqrt{n}\sigma$ yields
\[
\hat{\tau}^* = \frac{\lambda^*(\beta)}{\sqrt{\mu_\beta}}\, y_{\mathrm{med}}.
\]
For the square case $(\beta = 1)$, the MP median is $\mu_1 \approx 0.676$, giving
\[
\hat{\tau}^* \approx 2.858\, y_{\mathrm{med}}.
\]
Hence, the constant $2.858$ arises from the ratio between the theoretical optimal coefficient $\lambda^*(1) = 4/\sqrt{3}$ and the square root of the MP median $\sqrt{\mu_1}$. This empirical rule offers a convenient, parameter-free threshold that remains close to the theoretical optimum even when $\sigma$ is unknown.

In summary, the optimal hard threshold $\tau^*$ provides an elegant and theoretically grounded approach to low-rank matrix denoising. Its closed-form formula, absence of tuning parameters, and proven minimax optimality make it a cornerstone for modern, non-iterative matrix recovery methods based on SVHT.

\subsection{Proposed Method}
\subsubsection{Observation Model}

We consider the following noisy observation model for an $N$-way tensor:
\begin{equation*}
\label{eq:noise_model}
\mathcal{Y} = \mathcal{X} + \sigma \mathcal{E}, 
\quad \text{Rank}(\mathcal{X}) = \left(R_{I_1}, \dots, R_{I_N}\right), 
\quad 0 < R_{I_k} \le I_k,
\end{equation*}
where $\mathcal{Y} \in \mathbb{R}^{I_1 \times \cdots \times I_N}$ is the observed tensor, 
$\mathcal{X}$ is the true low-rank signal tensor, 
and $\mathcal{E}$ is an i.i.d. Gaussian noise tensor whose entries follow $\mathcal{N}(0,1)$. 
The scalar $\sigma > 0$ controls the standard deviation of the additive noise, 
i.e., each entry of $\mathcal{Y}$ satisfies 
$Y_{i_1,\dots,i_N} = X_{i_1,\dots,i_N} + \sigma \varepsilon_{i_1,\dots,i_N}$ 
with $\varepsilon_{i_1,\dots,i_N} \sim \mathcal{N}(0,1)$. 
The Tucker rank $\mathrm{Rank}(\mathcal{X})$ characterizes the multilinear rank of $\mathcal{X}$ 
and is defined through the ranks of the mode-$k$ unfoldings:
\[
R_{I_k} = \mathrm{Rank}\left(\mathbf{X}_{(k)}\right).
\]
Our objective is to estimate $\mathcal{X}$ from the noisy observation $\mathcal{Y}$.

\subsubsection{Mode-wise Denoising via SVHT}

Following the matrix denoising framework, we apply SVHT to each mode-$k$ unfolding of $\mathcal{Y}$. 
Let $\mathbf{Y}_{(k)},\mathbf{X}_{(k)}\in \mathbb{R}^{I_k \times (I_1 \cdots I_{k-1} I_{k+1} \cdots I_N)}$ denote the mode-$k$ unfoldings of $\mathcal{Y}$ and $\mathcal{X}$, respectively. 
Then, the mode-$k$ observation model is written as:
\begin{equation*}
\label{eq:matrix_noise_model}
\mathbf{Y}_{(k)} = \mathbf{X}_{(k)} + \sigma \mathbf{E}_{(k)},
\end{equation*}
where $\mathbf{E}_{(k)}$ is the mode-$k$ unfolding of the noise tensor $\mathcal{E}$. 

For each unfolding, we perform SVD:
\[
\mathbf{Y}_{(k)} = \sum_{i=1}^{d^*_{(k)}} y_{(k),i} \mathbf{u}_{(k),i} \mathbf{v}_{(k),i}^\top,
\]
where $d^*_{(k)} = \prod_{l \ne k} I_l$. 
Then, the denoised matrix $\hat{\mathbf{X}}_{(k)}$ is estimated using the hard thresholding rule:
\begin{equation*}
\label{eq:estimated_X}
\hat{\mathbf{X}}_{(k),\tau^*_{(k)}} 
= \sum_{i=1}^{d^*_{(k)}} 
\eta_H\left(y_{(k),i}; \tau^*_{(k)}\right) 
\mathbf{u}_{(k),i} \mathbf{v}_{(k),i}^\top, 
\end{equation*}

\begin{equation*}
\eta_H(y; \tau) =
\begin{cases}
y, & y \ge \tau, \\
0, & y < \tau.
\end{cases}
\end{equation*}
where the optimal threshold $\tau^*_{(k)}$ is adaptively determined as:
\[
\tau^*_{(k)} = \omega\left(\beta_{(k)}\right) \cdot y_{(k),\mathrm{med}}, \quad \beta_{(k)} = \frac{I_k}{d^*_{(k)}},
\]
with $\omega\left(\beta_{(k)}\right)$ denoting the aspect ratio-dependent scaling factor derived from the Optimal SVHT theory.

\subsubsection{Tensor Reconstruction}

After denoising each unfolding, we reconstruct the tensor using the dominant singular vectors that correspond to the retained singular values:
\[
\hat{\mathbf{U}}^{(k)} = \left[\mathbf{u}_{(k),1}, \dots, \mathbf{u}_{(k),g_{(k)}}\right], \quad
\hat{\mathbf{V}}^{(k)} = \left[\mathbf{v}_{(k),1}, \dots, \mathbf{v}_{(k),g_{(k)}}\right].
\]
The estimated Tucker core of the observed tensor is then computed as:
\[
\hat{\mathcal{S}}_y = \mathcal{Y} \times_1 \left(\hat{\mathbf{U}}^{(1)}\right)^\top \times_2 \cdots \times_N \left(\hat{\mathbf{U}}^{(N)}\right)^\top.
\]
Finally, the low-rank tensor estimate is reconstructed as:
\[
\hat{\mathcal{X}} = \hat{\mathcal{S}}_y \times_1 \hat{\mathbf{U}}^{(1)} \times_2 \cdots \times_N \hat{\mathbf{U}}^{(N)}.
\]

\subsubsection{Key Advantages and Theoretical Basis}

\begin{figure}[htbp]
    \centering
    \includegraphics[width=\textwidth]{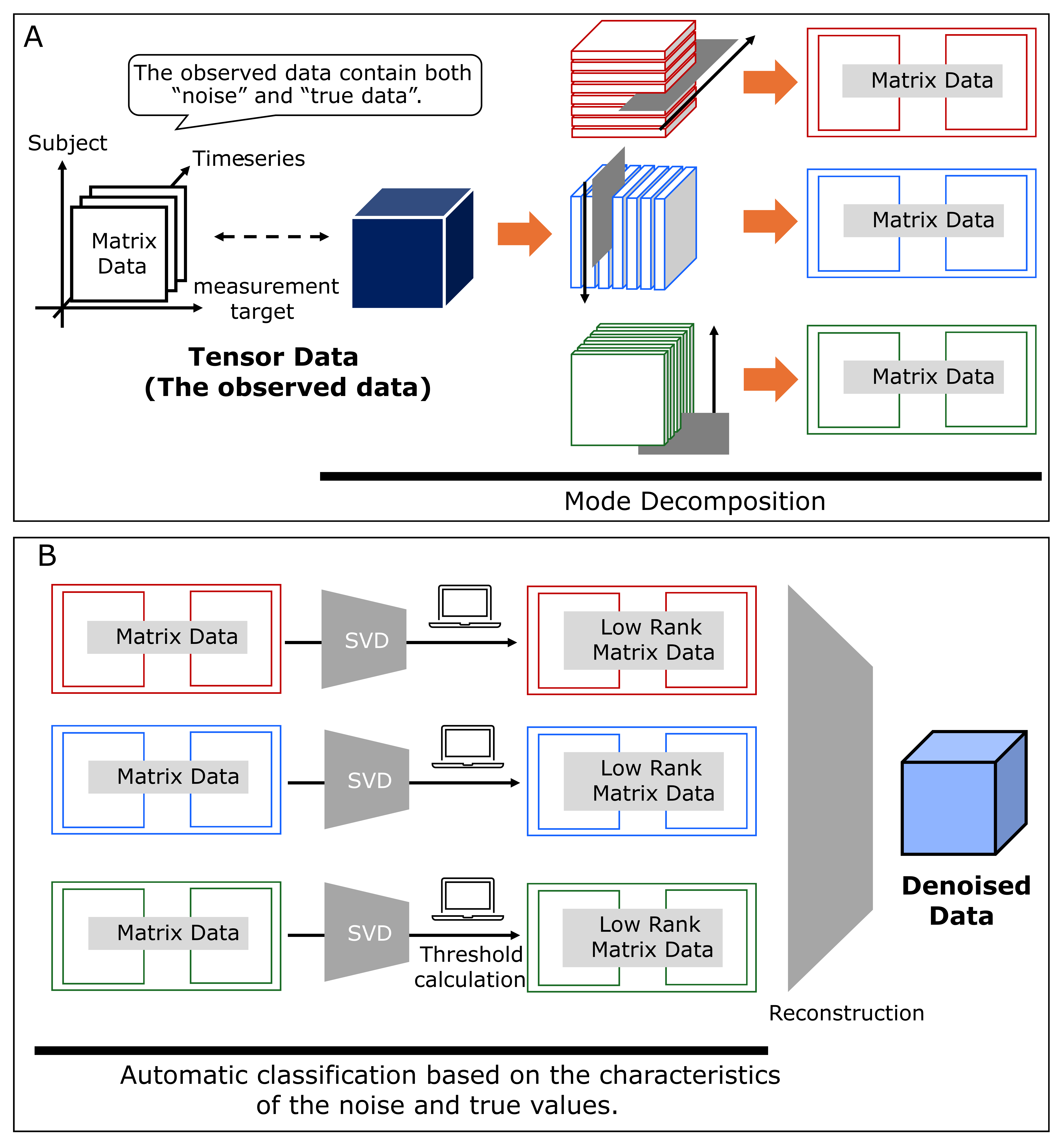}
    \caption{
    Overview of the proposed TARST method. (A) The observed tensor is decomposed mode-wise via SVD. (B) Statistically optimal thresholds are applied to remove noise, and the tensor is reconstructed to estimate the low-rank signal.
    Abbreviation: SVD, Singular Value Decomposition.}
    \label{fig:Overview}
\end{figure}

The proposed method, TARST, extends the Optimal SVHT framework from matrices to tensors. It automatically determines the rank of each mode through a statistically derived threshold, thereby eliminating the need for prior rank specification or iterative optimization as required in traditional HOSVD or HOOI. This property makes TARST a fully non-iterative and parameter-free tensor denoising algorithm. Furthermore, the overall workflow of TARST is illustrated in Fig.~\ref{fig:Overview} 
and summarized in Algorithm~\ref{alg:TARST}. 

\begin{algorithm}[H]
\caption{
TARST. 
The method denoises an observed tensor $\mathcal{Y}$ corrupted by additive noise via adaptive SVHT.
If the noise level $\sigma$ is known, the threshold is $\lambda^*(\beta)\sqrt{n}\sigma$; 
otherwise, it is estimated from the median singular value.
The denoised tensor is reconstructed by Tucker decomposition using the dominant singular vectors of each mode unfolding.
}
\label{alg:TARST}
\begin{algorithmic}[1]
\STATE \textbf{Input:} Tensor $\mathcal{Y}$, flag \texttt{sigma\_known}, noise level $\sigma$ (if known)
\STATE \textbf{Output:} Denoised tensor $\hat{\mathcal{X}}$
\STATE $N \gets$ number of modes of $\mathcal{Y}$; \quad $\mathbf{r} \gets [\,]$
\FOR{$k = 1$ to $N$}
    \STATE $\mathbf{Y}_{(k)} \gets$ mode-$k$ unfolding of $\mathcal{Y}$; \quad $(m, n) \gets$ shape of $\mathbf{Y}_{(k)}$
    \STATE $\beta \gets m/n$; \quad SVD: $\mathbf{Y}_{(k)} = \mathbf{U}_{(k)}\mathbf{S}_{(k)}\mathbf{V}_{(k)}^\top$
    \IF{\texttt{sigma\_known}}
        \STATE $\tau_{(k)} \gets \lambda^*(\beta)\sqrt{n}\sigma$
    \ELSE
        \STATE $\tau_{(k)} \gets \lambda^*(\beta)\,\mathrm{median}\left(\mathbf{S}_{(k)}\right)$
    \ENDIF
    \STATE $\hat{\mathbf{S}}_{(k)} \gets \mathrm{diag}\!\left(\max\{S_{(k),i}-\tau_{(k)},0\}\right)$
    \STATE $\hat{\mathbf{X}}_{(k)} \gets \mathbf{U}_{(k)}\hat{\mathbf{S}}_{(k)}\mathbf{V}_{(k)}^\top$
    \STATE Append $\mathrm{rank}\left(\hat{\mathbf{X}}_{(k)}\right)$ to $\mathbf{r}$
\ENDFOR
\STATE \textbf{Tucker reconstruction:}
\STATE $\hat{\mathcal{S}} \gets \mathcal{Y} \times_1 \mathbf{U}_{(1)}^\top \times_2 \cdots \times_N \mathbf{U}_{(N)}^\top$
\STATE $\hat{\mathcal{X}} \gets \hat{\mathcal{S}} \times_1 \mathbf{U}_{(1)} \times_2 \cdots \times_N \mathbf{U}_{(N)}$
\STATE \textbf{return} $\hat{\mathcal{X}}$
\end{algorithmic}
\end{algorithm}

\section{Simulations}
\subsection{Computational Cost Analysis}
This section presents a theoretical analysis of the computational complexity of the proposed TARST algorithm and derives its Worst-Case Time Ratio (WCTR).

Following standard asymptotic notation,
$\mathcal{O}(\cdot)$ and $\Omega(\cdot)$ denote the upper and lower bounds of computational complexity, respectively.
Formally, for two positive functions $f(n)$ and $g(n)$:
\[
\begin{aligned}
f(n) &= \mathcal{O}(g(n)) &&\text{if there exist constants } c>0,\, n_0>0 \text{ such that } f(n) \le c\,g(n) \text{ for all } n>n_0,\\[3pt]
f(n) &= \Omega(g(n)) &&\text{if there exist constants } c>0,\, n_0>0 \text{ such that } f(n) \ge c\,g(n) \text{ for all } n>n_0.\end{aligned}
\]
These notations describe the asymptotic behavior of computational cost with respect to the input size $P$ and mode dimensions $I_k$.

The analysis follows the worst-case complexity framework of \cite{Absi2019}, which evaluates algorithmic performance under the most unfavorable input conditions, and extends it from optimization to tensor decomposition.
Specifically, we derive the WCTR of TARST by analytically formulating the computational cost of each mode-wise operation and identifying the asymptotic conditions under which the theoretical upper bound becomes tight.

Let $\mathcal{Y} \in \mathbb{R}^{I_1 \times I_2 \times \cdots \times I_N}$ denote an input tensor, and let $P = \prod_{k=1}^{N} I_k$ represent its total number of elements.
The TARST algorithm consists of three primary steps.
First, for each mode $k$, the tensor is unfolded into a matrix $\mathbf{Y}_{(k)} \in \mathbb{R}^{I_k \times I_{(-k)}}$, where $I_{(-k)} = \prod_{j\neq k} I_j$, and SVD is performed.
Second, hard-thresholding is applied to the singular values using a threshold parameter $\tau_{(k)}$, which effectively suppresses small singular components.
Third, the tensor is reconstructed via mode-wise multiplications using the left singular matrices $\mathbf{U}_{(k)}$, where the post-threshold rank is defined as $r_k = \mathrm{rank}\left(\hat{\mathbf{X}}_{(k)}\right)$.
Under the dense linear algebra model, the information-theoretic lower bound is $T_{\mathrm{opt}}(\mathcal{Y}) = \Omega(P)$, since each input element must be read at least once.

The WCTR is defined as
\[
\mathrm{WCTR}(\text{method})
=
\sup_{\mathcal{Y}}
\frac{T_{\mathrm{method}}(\mathcal{Y})}{T_{\mathrm{opt}}(\mathcal{Y})},
\]
which quantifies the maximum deviation of the algorithm’s computational cost from the theoretical lower bound.
By evaluating this ratio for TARST, the theoretical existence, tightness, and asymptotic form of the computational upper bound can be determined.

The evaluation focuses on the derivation of the theoretical upper bound and the investigation of the existence conditions for a finite polynomial bound of TARST. Specifically, the theoretical upper bound of TARST’s computational cost is derived by decomposing and summing its mode-wise operations under the worst-case complexity framework \cite{Absi2019}, thereby establishing its asymptotic order. Furthermore, the analytical conditions under which this bound remains finite and polynomial are examined to ensure the theoretical soundness and computational stability of the proposed method. The results section summarizes these analyses, providing a comprehensive understanding of TARST’s computational characteristics and upper-bound behavior.

\subsection{Reconstruction Accuracy Analysis}
The objective of this analysis is to quantitatively evaluate the denoising and reconstruction performance of the proposed TARST method in comparison with existing tensor decomposition approaches. Using synthetic datasets with known ground-truth structures allows direct assessment of how accurately each method reconstructs the underlying signal under different noise and outlier conditions. All experiments were conducted in MATLAB R2025b using Tensor Toolbox v3.6 and the official TuckerL2E-master implementation available on GitHub. Default Tucker-L2E parameters were used, and the random seed was fixed (\texttt{rng(0)}) to ensure reproducibility.

The true tensors were designed to have predefined low-rank structures, which served as the ground-truth ranks in the experiments. For a fair comparison, rank-specified methods (HOSVD, HOOI) were provided with the correct rank information corresponding to the true low-rank structure, while the proposed TARST automatically estimated the rank for each mode without prior specification. Five methods were compared in total: Baseline, HOSVD, HOOI, Tucker-L2E, and the proposed TARST. The Baseline simply uses the noisy observation tensor as a reference for the effect of noise. HOSVD~\citep{DeLathauwer2000a} performs Tucker decomposition with a fixed rank in a non-iterative manner, while HOOI~\citep{DeLathauwer2000b} refines the HOSVD result iteratively through Alternating Least Squares (ALS) optimization. Tucker-L2E~\citep{Chen2020} is a robust extension of the Tucker model based on the $L_{2}E$ estimator, designed to mitigate the influence of outliers. Finally, the proposed TARST performs automatic rank estimation and singular value hard thresholding for each mode without iterative optimization, enabling fast and parameter-free tensor reconstruction.

Two types of synthetic experiments were conducted to evaluate performance under different noise and outlier conditions. In the first experiment (Pattern~1), Gaussian noise was added to assess denoising capability. Two tensor sizes were used: $10 \times 10 \times 10$ and $50 \times 50 \times 50$. Each tensor was generated from a normal distribution with mean 10 and standard deviations of either 2 or 0.25, resulting in four total combinations. Gaussian noise with standard deviation $\sigma \in [10^{-1}, 10^{1}]$ (log scale) was added as $\mathcal{Y} = \mathcal{X} + \sigma \mathcal{E}$, where $\mathcal{Y}$ is the observed tensor, $\mathcal{X}$ is the ground truth, and $\mathcal{E} \sim \mathcal{N}(0,1)$ is Gaussian noise. While $\sigma$ is known in these synthetic experiments to control the noise level, the proposed TARST does not require prior knowledge of $\sigma$ during decomposition, as the threshold is automatically determined from the empirical singular-value distribution. A smaller $\sigma$ corresponds to low-noise conditions, whereas a larger $\sigma$ represents high-noise conditions. Each condition was repeated five times, and the mean relative reconstruction squared error (RRSE) with 95\% confidence intervals was reported. 
The RRSE~\citep{Zhao2015} is defined as
\[
\mathrm{RRSE}(\hat{\mathcal{X}}, \mathcal{X})
= \frac{\|\hat{\mathcal{X}} - \mathcal{X}\|_F}{\|\mathcal{X}\|_F},
\]
where $\hat{\mathcal{X}}$ denotes the reconstructed tensor and $\mathcal{X}$ the ground truth. 
A smaller RRSE indicates a more accurate recovery of the true underlying tensor structure.

In the second experiment (Pattern~2), robustness to outliers was examined. A $10 \times 10 \times 10$ tensor with elements drawn from $\mathcal{N}(10, 2^2)$ was used as the ground truth. Gaussian noise with $\sigma \in [0.1, 10]$ was added, and outliers were introduced by replacing randomly selected elements with values scaled by $\{10, 25, 50, 100\}$. The outlier ratio was varied across $\{1\%, 5\%, 10\%, 25\%, 50\%\}$, with each setting repeated five times. Results were visualized as RRSE heatmaps, where the horizontal and vertical axes represent the noise level and outlier scale, respectively. The color scale was unified across all subplots for consistent comparison of RRSE magnitudes among the methods.

\section{Results}

\subsection{Computational Cost Analysis}
\subsubsection{Derivation of the Theoretical Upper Bound}
The computational cost of the proposed TARST algorithm can be divided into three major phases for each tensor mode: (1) SVD of the mode-unfolded matrices, (2) hard thresholding of singular values, and (3) Tucker-type reconstruction using the post-threshold components.
For a tensor with $N$ modes, the overall computational cost is expressed as

\[
T_{\mathrm{TARST}}
=
\sum_{k=1}^{N}
\left(
T_{\mathrm{SVD},k}
+
T_{\mathrm{thresh},k}
+
T_{\mathrm{recon},k}
\right).
\]
Each phase can be asymptotically formulated as follows:
\[
T_{\mathrm{SVD},k} = \mathcal{O}\!\left(\min\left\{I_k^2 I_{(-k)},\, I_k I_{(-k)}^2\right\}\right),
\quad
T_{\mathrm{thresh},k} = \mathcal{O}\!\left(\min\left\{I_k, I_{(-k)}\right\}\right),
\quad
T_{\mathrm{recon},k} = \mathcal{O}\!\left(\frac{P\, r_{(-k)}}{I_k}\right),
\]
where $I_{(-k)} = \prod_{j \neq k} I_j$ and $r_{(-k)} = \prod_{j \neq k} r_j$.

By aggregating these three components, the overall theoretical upper bound of TARST is given by

\[
T_{\mathrm{TARST}}
=
\mathcal{O}\!\Bigg(
\underbrace{\sum_{k=1}^{N} \min\left\{I_k^2 I_{(-k)}, I_k I_{(-k)}^2\right\}}_{\text{(1) SVD phase}}
+
\underbrace{\sum_{k=1}^{N} \min\left\{I_k, I_{(-k)}\right\}}_{\text{(2) hard-thresholding phase}}
+
\underbrace{\sum_{k=1}^{N} \frac{P\, r_{(-k)}}{I_k}}_{\text{(3) Reconstruction phase}}
\Bigg).
\]

The first term corresponds to the SVD operations performed on each mode-unfolded matrix,  and it dominates the computational cost when the tensor dimensions $I_k$ are large.  The second term represents the cost of hard-thresholding the singular values,  which increases linearly with the number of singular values $\min\left\{I_k, I_{(-k)}\right\}$.  Although this cost is asymptotically smaller than that of the SVD,  it is a constant and non-negligible overhead across all modes.  The third term represents the cost of the Tucker-type reconstruction,  which depends on the post-threshold ranks $r_k$ and becomes dominant in low-rank conditions.  When $r_k \ll I_k$, the reconstruction term dominates and the total cost approaches linear order in $P$.  In contrast, under the worst-case scenario where $r_k = I_k$,  the complexity converges to the same order as HOSVD.

Consequently, the total theoretical upper bound of TARST can be summarized as

\begin{equation}
T_{\mathrm{TARST}}
=
\mathcal{O}\!\Bigg(
\sum_{k=1}^{N}
\Bigg[
\min\left\{I_k^2 I_{(-k)},\, I_k I_{(-k)}^2\right\}
+
\min\left\{I_k,\, I_{(-k)}\right\}
+
\frac{P\, r_{(-k)}}{I_k}
\Bigg]
\Bigg).
\label{eq:upperbound_general}
\end{equation}

This result theoretically guarantees that the computational cost of TARST is upper-bounded  as a function of the mode dimensions $I_k$, the total number of tensor elements $P$,  and the post-threshold ranks $r_k$. Therefore, TARST operates stably within polynomial time in the high-rank regime and achieves significant computational reduction under low-rank conditions.
\subsubsection{Polynomial-Time Bound under General Conditions}
\label{sec:general-complexity}

We next derive the theoretical upper bound of TARST under general, anisotropic conditions without assuming equal tensor mode sizes. 
Let the tensor have $N < \infty$ modes, and denote its mode dimensions and post-threshold ranks as $I_k$ and $r_k$, respectively. Assume that the ranks increase more slowly than their corresponding mode dimensions, i.e.,
\[
r_k = \mathcal{O}(I_k^{\alpha_k}), 
\qquad 0 \le \alpha_k < 1, 
\quad k = 1, 2, \dots, N.
\]
Let $P = \prod_{k=1}^{N} I_k$ denote the total number of tensor elements. Then, the following proposition holds.

\begin{proposition}[Polynomial-Time Upper Bound under General Anisotropic Conditions]
\label{prop:poly-upper-bound}
Under the above assumptions, the total computational cost of TARST satisfies
\[
T_{\mathrm{TARST}} = \mathcal{O}(P^{1+\varepsilon}),
\]
where $\varepsilon = \max\{1/2,\, \alpha_{\max}\}$ and $\alpha_{\max} = \max_k (\alpha_k) < 1$.  Hence, TARST operates within polynomial time with respect to the input size $P$.
\end{proposition}

\textbf{Proof.}
The computational upper bound in Eq.~\eqref{eq:upperbound_general} consists of three major terms:  the SVD term, the hard-thresholding term, and the reconstruction term.  Each term is now analyzed separately.

\textit{(1) SVD term.}  
The asymptotic cost of the SVD for each mode is 
\[
T_{\mathrm{SVD},k} = \mathcal{O}\!\left(\min\left\{I_k^2 I_{(-k)},\, I_k I_{(-k)}^2\right\}\right).
\]
Because $P = \prod_{j=1}^{N} I_j = I_k I_{(-k)}$,  this expression can be rewritten in terms of $I_k$ and $P$ as 
\[
\min\left\{I_k P,\, \frac{P^2}{I_k}\right\}.
\]
By applying the arithmetic–geometric mean inequality, which holds since all elements are positive, we can set a loose but valid upper bound as
\[
\min\left\{I_k P,\, \frac{P^2}{I_k}\right\} \leq \sqrt{(I_k P)\!\left(\frac{P^2}{I_k}\right)} = P^{3/2}.
\]
Thus, the SVD term is upper-bounded by $\mathcal{O}(P^{3/2})$.

\textit{(2) Hard-thresholding term.}  
The thresholding phase has asymptotic cost
\[
T_{\mathrm{thresh},k} = \mathcal{O}\!\left(\min\{I_k, I_{(-k)}\}\right)
= \mathcal{O}\!\left(\min\left\{I_k, \frac{P}{I_k}\right\}\right).
\]
Similarly, using the same bounding technique,
\[
\min\left\{I_k, \frac{P}{I_k}\right\} \leq \sqrt{I_k \left(\frac{P}{I_k}\right)} = P^{1/2}.
\]
Hence, the hard-thresholding term contributes an order of $\mathcal{O}(P^{1/2})$,  which is asymptotically smaller than the SVD term.

\textit{(3) Reconstruction term.}  
The reconstruction phase is expressed as
\[
T_{\mathrm{recon},k} = \mathcal{O}\!\left(\frac{P\, r_{(-k)}}{I_k}\right).
\]
To evaluate $r_{(-k)}$, we assume $r_j = \mathcal{O}(I_j^{\alpha_j})$ for each mode,  where $0 \le \alpha_j < 1$.  Let $\alpha_{\max} = \max_j (\alpha_j)$.  Then, for any $j$, we have $r_j \leq C_j I_j^{\alpha_{\max}}$,  where $C_j$ is a constant depending only on mode $j$.  Therefore, 
\[
r_{(-k)} = \prod_{j \neq k} r_j 
\leq \left(\prod_{j \neq k} C_j\right) 
\left(\prod_{j \neq k} I_j^{\alpha_{\max}}\right)
= C \cdot I_{(-k)}^{\alpha_{\max}},
\]
where $C = \prod_{j \neq k} C_j$ is a constant.  Since $I_{(-k)} = P / I_k$,  this becomes $r_{(-k)} \leq C (P / I_k)^{\alpha_{\max}}$.  Substituting this into the reconstruction term yields
\[
\frac{P\, r_{(-k)}}{I_k}
\leq C \cdot \frac{P}{I_k} \left(\frac{P}{I_k}\right)^{\alpha_{\max}}
= C \cdot P^{1+\alpha_{\max}} I_k^{-(1+\alpha_{\max})}.
\]
Because $I_k \ge 1$, the inverse power term $I_k^{-(1+\alpha_{\max})}$  is bounded above by one, resulting in
\[
T_{\mathrm{recon},k} = \mathcal{O}(P^{1+\alpha_{\max}}).
\]

\textit{(4) Summary.}  
Combining the three terms, we obtain
\[
T_{\mathrm{TARST}} = 
\mathcal{O}\!\left(
P^{3/2} + P^{1/2} + P^{1+\alpha_{\max}}
\right).
\]
Among these, the dominant exponent is 
$\max\{3/2, 1+\alpha_{\max}\} = 1 + \max\{1/2, \alpha_{\max}\}$.  
Letting $\varepsilon = \max\{1/2, \alpha_{\max}\}$,  and noting that $\alpha_{\max} < 1$,  we conclude that 
\[
T_{\mathrm{TARST}} = \mathcal{O}(P^{1+\varepsilon}), 
\qquad \varepsilon < 1.
\]
Therefore, the theoretical computational complexity of TARST is bounded within polynomial time, and the exponent is strictly less than two.  
\hfill $\square$

\begin{corollary}[Boundedness of the Worst-Case Time Ratio]
Let the optimal computational lower bound be $T_{\mathrm{opt}} = \Omega(P)$.  Then, from the above proposition, the WCTR of TARST satisfies
\[
\mathrm{WCTR}(\mathrm{TARST}) = 
\mathcal{O}(P^{\varepsilon}),
\qquad
\varepsilon < 1.
\]
This result indicates that the WCTR of TARST always remains finite,  and the computational growth rate is strictly linear with respect to the input size $P$.  Therefore, TARST guarantees stable computational scalability within polynomial time,  even under general anisotropic settings.
\end{corollary}

When the isotropic condition ($I_k = I$, $r_k = r$) is additionally assumed,  the general expression derived above simplifies considerably.  Because the constant multiplicative factor can be separated from the asymptotic order  (and thus $N$ can be omitted),  the computational cost reduces to
\[
T_{\mathrm{TARST}}
=
\mathcal{O}\!\Big(
I^{N+1} + I + I^{N-1} r^{N-1}
\Big).
\]
In this formulation, the first term $I^{N+1}$ corresponds to the SVD cost for each mode,  the second term $I$ represents the hard-thresholding step,  and the third term $I^{N-1} r^{N-1}$ denotes the reconstruction cost.  If we further assume that $r = \mathcal{O}(I^\alpha)$ with $0 \leq \alpha < 1$,  the overall computational cost becomes
\[
T_{\mathrm{TARST}} = \mathcal{O}(P^{1+\frac{1}{N}}),
\]
which is consistent with the theoretical bound obtained under the general anisotropic setting.  This isotropic formulation provides an intuitive interpretation of the asymptotic scaling behavior of TARST: the dominant term $I^{N+1}$ grows polynomially with respect to the tensor dimension,  and the exponent $1+\tfrac{1}{N}$ becomes smaller as the tensor order $N$ increases,  indicating that higher-order tensors enhance the relative computational stability of TARST.

\subsubsection{Existence Conditions for the Polynomial-Time Bound}
\label{sec:polynomial-conditions}

Based on the polynomial-time upper bound derived in Proposition~\ref{prop:poly-upper-bound}, we now clarify the theoretical conditions under which a finite polynomial-time bound exists for TARST.

For a dense tensor $\mathcal{Y} \in \mathbb{R}^{I_1 \times \cdots \times I_N}$ with $P = \prod_{k=1}^{N} I_k$ elements, let $r_k = \mathcal{O}(I_k^{\alpha_k})$ denote the post-threshold rank in each mode. Then, the computational cost of TARST satisfies 
\[
T_{\mathrm{TARST}} = \mathcal{O}(P^{1+\varepsilon}), \qquad \varepsilon < 1,
\]
provided that the following two general conditions hold:

\begin{itemize}
    \item[(C1)] The rank-growth exponents satisfy $\alpha_k < 1$ for all $k = 1, 2, \dots, N$, 
    ensuring that each mode-wise rank increases more slowly than its corresponding mode dimension.  
    This prevents the reconstruction term from dominating the asymptotic order.
    
    \item[(C2)] The tensor order $N$ is finite ($N < \infty$), 
    so that the mode-wise summation in Eq.~\eqref{eq:upperbound_general} 
    contributes only a constant multiplicative factor to the total cost.
\end{itemize}

Under these assumptions, the exponent $\varepsilon = \max\{1/2, \alpha_{\max}\}$ with $\alpha_{\max} = \max_k (\alpha_k)$ remains strictly less than one. Consequently, TARST is guaranteed to operate within polynomial time, and its computational complexity exponent is strictly smaller than two.  This ensures that the algorithm achieves stable computational scalability even for high-order or anisotropic tensor data.

\subsection{Reconstruction Accuracy Analysis}
\subsubsection{Pattern 1 (Gaussian noise)}

\begin{figure}[ht]
    \centering
    \includegraphics[width=\textwidth]{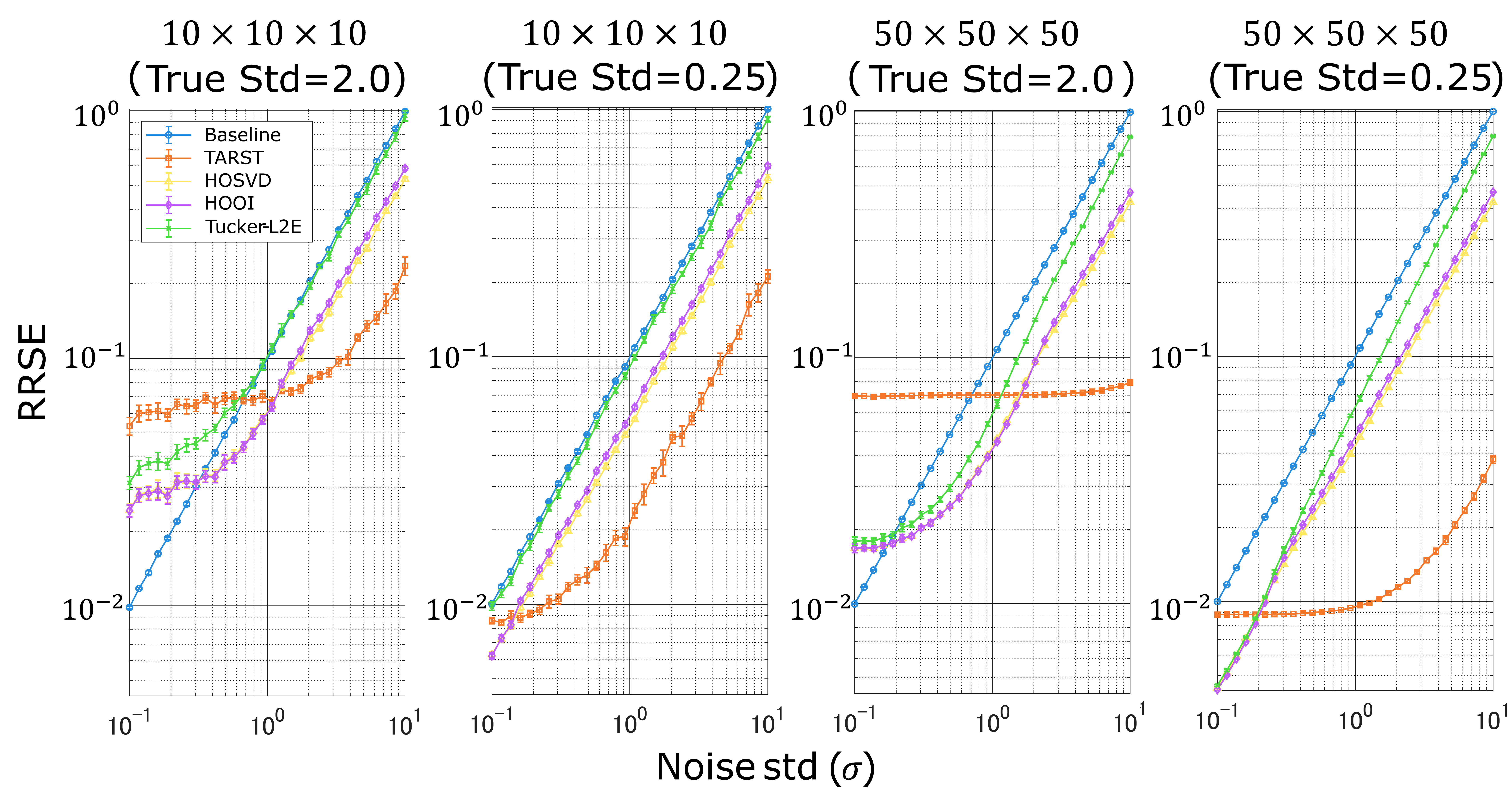}
    \caption{
    Results of Pattern 1 (Gaussian Noise). 
    From left to right, the figures correspond to tensors of size $10\times10\times10$ with true tensor mean $10$ and standard deviation $2$, 
    $10\times10\times10$ with mean $10$ and standard deviation $0.25$, 
    $50\times50\times50$ with mean $10$ and standard deviation $2$, 
    and $50\times50\times50$ with mean $10$ and standard deviation $0.25$. 
    Five methods are compared: Baseline (no denoising), HOSVD, HOOI, Tucker-L2E, and TARST. 
    Both axes are shown on a logarithmic scale: 
    the horizontal axis represents the noise level $\sigma$ (standard deviation of additive Gaussian noise),  and the vertical axis shows the RRSE, where smaller values indicate better performance.
    }
    \label{fig:Result2}
\end{figure}

Figure~\ref{fig:Result2} illustrates how the relative RRSE varies with the noise standard deviation $\sigma$ across four experimental settings (small/large scale × low/high variance).
Here, $\sigma$ represents the noise level that scales additive Gaussian noise generated from $\mathcal{N}(0,1)$, resulting in zero-mean noise with variance $\sigma^2$.

Overall, RRSE increases monotonically with the noise level for all methods. TARST consistently achieves the lowest RRSE among the compared approaches and demonstrates only gradual error growth even under high-noise conditions ($\sigma > 1$). In contrast, both HOSVD and HOOI exhibit a sharp increase in RRSE once $\sigma$ exceeds 0.3, indicating substantial degradation in reconstruction accuracy at higher noise levels. Tucker-L2E performs favorably in the low-noise region ($\sigma < 0.3$) but deteriorates rapidly thereafter. The baseline method produces the highest RRSE values throughout, with errors increasing almost proportionally to the noise magnitude.

In the small-scale experiments ($10\times10\times10$), noise effects are particularly pronounced. When the underlying signal variance is large (Std = 2.0), RRSE values for HOSVD, HOOI, and Tucker-L2E rise steeply as $\sigma$ increases. Even under the low-variance condition (Std = 0.25), a noticeable jump in error appears once $\sigma$ exceeds 1. TARST maintains lower RRSE than the other methods across all noise levels, and its error growth remains roughly linear with respect to the noise magnitude, except in the large-scale, high-variance setting ($50\times50\times50$, True Std = 2.0).

For larger tensors ($50\times50\times50$), RRSE values generally decrease for all methods, suggesting a stabilizing effect of data scale. Although HOSVD and HOOI continue to show increasing errors as $\sigma$ rises, TARST consistently maintains low RRSE across the entire range of noise levels. In the large-scale, low-variance condition, all methods yield relatively small RRSE overall, but both HOSVD and HOOI exhibit a noticeable increase when $\sigma$ exceeds 1. Across all four experimental conditions, RRSE tends to grow with noise intensity, while TARST consistently achieves the highest robustness and lowest reconstruction error.

\subsubsection{Pattern 2 (Outlier robustness)}

\begin{figure}[ht]
    \centering
    \includegraphics[width=\textwidth]{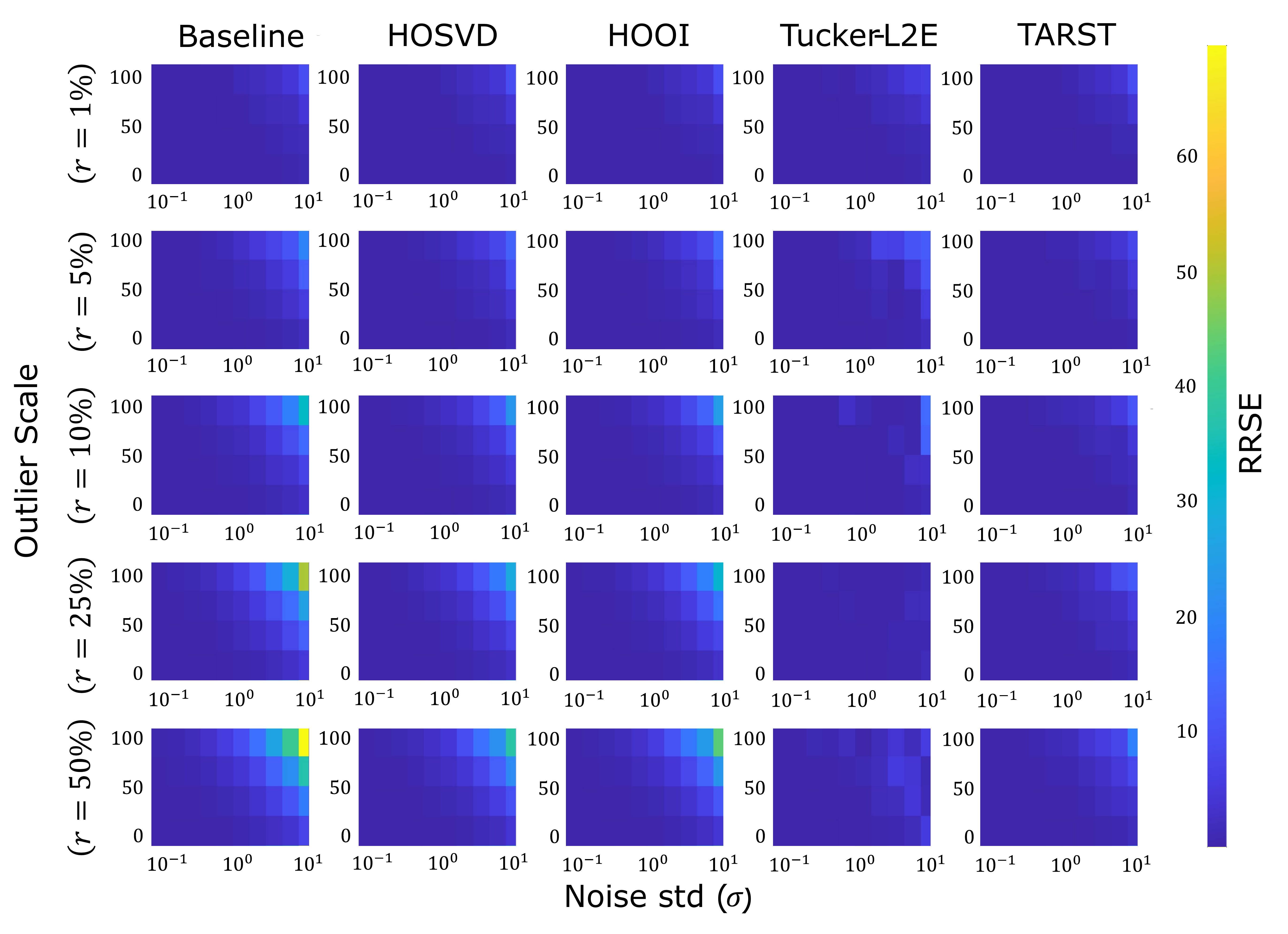}
    \caption{
Results of Pattern 2 (Outlier Robustness).This figure illustrates how each method (Baseline, HOSVD, HOOI, Tucker-L2E, TARST) responds to changes in outlier ratio and noise dispersion, visualized as a heatmap of RRSE values.The vertical axis represents the outlier ratio $r$. For example, $r=10\%, 50$ indicates that 10\% of true values are replaced with outliers scaled by a factor of 50. The horizontal axis denotes the noise level, corresponding to the standard deviation of Gaussian noise, ranging from $10^{-1}$ to $10^{1}$.
The color scale shows the RRSE (Relative Reconstruction Squared Error): smaller values (blue) indicate better performance, while larger values (yellow) denote higher error.
    }
    \label{fig:Result3}
\end{figure}

Figure~\ref{fig:Result3} presents the RRSE obtained under combined variations of noise level ($\sigma$) and outlier conditions for five tensor decomposition methods: Baseline, HOSVD, HOOI, Tucker-L2E, and TARST.
Each column corresponds to a method, while each row represents a different outlier ratio $r$ ($1\%, 5\%, 10\%, 25\%, 50\%$).
The horizontal axis shows the standard deviation of additive Gaussian noise ($\sigma$), ranging from $10^{-1}$ to $10^{1}$ on a logarithmic scale. Here, $\sigma$ represents the noise level that scales additive Gaussian noise generated from $\mathcal{N}(0,1)$, resulting in zero-mean noise with variance $\sigma^2$. The vertical axis within each heatmap represents the outlier scale, which indicates the magnitude by which a portion of true data points—specified by $r$—are amplified.
For example, $r = 5\%, 50$ denotes a condition in which $5\%$ of the true entries are replaced by values that are 50 times larger than their original magnitudes. Color intensity represents RRSE, where darker (blue) regions indicate lower reconstruction error and lighter (yellow) regions correspond to higher error.

Across all noise and outlier conditions, the Baseline method exhibits high RRSE throughout, reflecting large reconstruction error even for small noise levels. HOSVD and HOOI achieve lower RRSE under low noise and small amplification scales but show progressively lighter regions as $\sigma$ or the outlier scale increases, indicating reduced reconstruction accuracy. Tucker-L2E maintains moderate error levels at small $r$ and low amplification factors but displays higher RRSE when the outlier scale exceeds about 25× or when $\sigma > 1$. TARST yields consistently dark regions across all panels, suggesting uniformly low RRSE irrespective of noise intensity, outlier ratio, or outlier scale. Even under the most severe conditions ($r=50\%$, outlier scale = 100, $\sigma=10$), RRSE variation for TARST remains minimal, showing the most stable reconstruction among the compared methods.

\section{Discussion}

The theoretical analysis confirmed that TARST achieves tractable computational scalability, operating within polynomial time $\mathcal{O}(P^{1+\varepsilon})$ under two general conditions: (i) the post-threshold ranks of each mode grow more slowly than their corresponding mode dimensions ($\alpha_k < 1$), and (ii) the tensor order $N$ is finite. Under these assumptions, the exponent $\varepsilon = \max{1/2, \alpha_{\max}}$ remains strictly below one, indicating that the computational complexity exponent is smaller than two. This result theoretically guarantees stable scalability even for high-order or anisotropic tensor data and explains why the proposed method maintains tractable computation as the tensor dimension or data volume increases. In addition to computational efficiency, TARST generalizes the optimal singular-value thresholding principle of the MP law to the tensor domain, providing the first statistically grounded framework for automatic rank determination in tensor decomposition. Moreover, by incorporating the WCTR analysis framework~\cite{Absi2019}, this study introduces a new theoretical perspective for assessing algorithmic efficiency in tensor computation, bridging statistical optimality with computational tractability in OR.

Experimentally, TARST consistently achieved the lowest RRSE across all settings and maintained stable performance under increasing noise levels and outlier ratios. This robustness arises from its automatic thresholding mechanism based on the MP distribution, which dynamically adjusts the threshold according to noise intensity, removing noise-dominated singular values while preserving true signal components. Consequently, error growth remains nearly linear with respect to $\sigma$, and higher-dimensional settings further enhance stability due to the concentration of the singular-value distribution around its expectation. Because TARST automatically determines the statistically valid rank for each mode, it avoids the under- and overestimation errors frequently observed in fixed-rank methods such as HOSVD and HOOI. Moreover, its non-iterative structure removes dependence on initialization and local minima, ensuring both computational efficiency and stability. Overall, by combining automatic rank estimation with a non-iterative design, TARST mitigates the bias of fixed-rank methods and the variance of iterative ones, 
maintaining consistent robustness against noise and scale variation.

In the outlier experiments (Figure~\ref{fig:Result3}), RRSE surfaces reveal that HOSVD and HOOI exhibit substantial error increases as the outlier ratio or outlier scale increases. In contrast, Tucker-L2E demonstrates partial robustness, maintaining low RRSE under moderate contamination and achieving lower errors than other methods around $r = 20\%$, where its loss function mitigates the impact of moderate outliers. However, as contamination increases (e.g., $r>25\%$), its performance deteriorates. 
TARST, by comparison, maintains uniformly low RRSE values across nearly all conditions, showing minimal variation even under extreme settings (e.g., $r=50\%$, scale = 100). This indicates that the proposed thresholding strategy limits the effect of extreme observations by discarding outlier-dominated singular components, resulting in stable reconstruction performance across contamination conditions. These results suggest that TARST adapts well to situations where a small fraction of data exhibits highly dispersed values—analogous to heavy-tailed noise found in real-world datasets such as financial transactions, energy demand, or sensor readings. Unlike fixed-rank or iterative approaches, TARST automatically adjusts its rank to exclude such fluctuations without relying on an explicit noise model. This behavior implies that MP-based thresholding acts as a robustness regularizer, suppressing extreme singular components while preserving the global multilinear structure. Consequently, TARST demonstrates both noise resistance and resilience to heavy-tailed perturbations, bridging statistical robustness with computational tractability in tensor decomposition.

Practically, TARST’s non-iterative and parameter-free design ensures reproducible results independent of analyst expertise or initialization. Its robustness to noise and outliers makes it suitable for various OR problems, including demand forecasting, anomaly detection, and efficiency evaluation involving multidimensional data. Furthermore, its simplicity and scalability enable deployment in large-scale, real-time analytical environments such as IoT and smart city monitoring.

A current limitation of TARST is its assumption of i.i.d. Gaussian noise. Future research should extend the framework to handle non-Gaussian, anisotropic, or time-varying noise. Algorithmically, developing distributed or online implementations and incorporating adaptive mode-wise thresholding would improve scalability for streaming and large-scale data. From an application viewpoint, applying TARST to real-world datasets in domains such as transportation, finance, and energy, and integrating it with optimization-based decision-support systems represent promising directions. Additionally, establishing asymptotic guarantees for the estimated thresholds and comparing TARST’s statistical efficiency with other robust estimators would strengthen its theoretical foundation. In summary, TARST achieves theoretical optimality and computational efficiency while overcoming key limitations of conventional tensor decomposition methods, including rank specification, iterative optimization, and initialization sensitivity. The proposed framework provides a robust, reproducible, and theoretically grounded foundation for high-dimensional data analysis, advancing data-driven decision support in OR.

\section{Conclusion}
This study proposed TARST, a non-iterative and rank-free tensor decomposition method designed to overcome two major limitations of conventional approaches: the requirement for pre-specified ranks and the computational instability inherent in iterative optimization. By employing automatic thresholding based on the MP distribution, TARST achieves statistically valid rank estimation. Theoretical analysis using the WCTR further demonstrated that TARST guarantees computational tractability within polynomial time. Accordingly, TARST provides a unified methodological foundation for tensor decomposition that ensures both theoretical optimality and computational efficiency.

Simulation experiments demonstrated that TARST maintained the lowest reconstruction error across diverse noise intensities and outlier ratios, exhibiting remarkable stability compared with existing methods. These results indicate that the proposed approach is consistently robust to noise, outliers, and scale variation. In particular, its ability to estimate the underlying structure without rank specification or iterative refinement offers significant practical advantages.

From the perspective of OR, TARST contributes to both methodological innovation and practical decision support. Methodologically, it establishes a statistically grounded and computationally tractable framework for high-dimensional tensor analysis, bridging the gap between statistical estimation theory and algorithmic efficiency in OR. Practically, the robustness and parameter-free nature of TARST enable reliable, analyst-independent analysis in data-driven decision-support contexts such as demand forecasting, anomaly detection, and performance evaluation. Thus, TARST serves as a foundational analytical tool that connects theoretical rigor with operational applicability.

Future research directions include extending the framework to accommodate non-Gaussian and time-varying noise, developing distributed or online implementations, and integrating TARST with optimization algorithms to support real-time decision-making. Through these extensions, TARST has the potential to further bridge theory and practice in high-dimensional data analysis and contribute to the continued advancement of OR.

\section{Acknowledgements}
This work was supported by JSPS KAKENHI Grant Number 23K22166.

\section{Declaration of generative AI and AI-assisted technologies}
During the preparation of this work, the authors used ChatGPT-4o in order to improve the quality of English translation. After using this tool, the authors reviewed and edited the content as needed and take full responsibility for the content of the published article.

\end{document}